\def\year{2018}\relax
\documentclass[letterpaper]{article} 
\usepackage{aaai18}  
\usepackage{times}  
\usepackage{helvet}  
\usepackage{courier}  
\usepackage{url}  
\usepackage{graphicx}  
\usepackage{subfigure}
\usepackage{amsmath}
\usepackage{array}
\usepackage{algorithm}
\usepackage{algorithmic}
\usepackage[american]{babel}
\begin{document}
%
\title{Deep Neural Network Compression with Single and Multiple Level Quantization}
\author{Yuhui Xu$^1$\thanks{Corresponding authors are Weiyao Lin and Hongkai Xiong: \{wylin, xionghongkai\}@sjtu.edu.cn.}, Yongzhuang Wang$^1$, Aojun Zhou$^2$, Weiyao Lin$^1$, Hongkai Xiong$^1$ \\
 $^1$ School of Electronic Information and Electrical Engineering, Shanghai Jiao Tong University, China\\
 $^2$ University of Chinese Academy of Sciences, China \\
 }

\maketitle
\begin{abstract}
Network quantization is an effective solution to compress deep neural networks for practical usage. Existing network quantization methods cannot sufficiently exploit the depth information to generate low-bit compressed network. In this paper, we propose two novel network quantization approaches, single-level network quantization (SLQ) for high-bit quantization and multi-level network quantization (MLQ) for extremely low-bit quantization (ternary). We are the first to consider the network quantization both from width and depth level. In the width level, parameters are divided into two parts: one for quantization and the other for re-training to eliminate the quantization loss. SLQ leverages the distribution of the parameters to improve the width level. In the depth level, we introduce incremental layer compensation to quantize layers iteratively which decreases the quantization loss in each iteration. The proposed approaches are validated with extensive experiments based on the state-of-the-art neural networks including AlexNet, VGG-16, GoogleNet and ResNet-18. Both SLQ and MLQ achieve impressive results. Code is available at \url{https://github.com/yuhuixu1993/SLQ}.
\end{abstract}

\section{Introduction}
Recent years, deep convolutional neural networks (DNNs) are playing an important role in a variety of computer vision tasks including image classification \cite{Krizhevsky2012ImageNet}, object detection \cite{Girshick_2015_ICCV,ren2015faster}, semantic segmentation \cite{chen2014semantic,Long_2015_CVPR} and face recognition \cite{Taigman_2014_CVPR,Sun_2014_CVPR}. The promising results of DNNs are contributed by many factors. Regardless of more training resources and powerful computational hardware, the large number of learnable parameters is the most important one. To achieve high accuracy, deeper and wider networks are designed which in turn poses heavy burden on storage and computational resources. It becomes more difficult to deploy a typical DNN model on resource constrained mobile devices such as mobile phones and drones. Thus, network compression is critical and has become an effective solution to reduce the storage and computation costs for DNN models.

\begin{figure}
  \centering
  \subfigure[Single-level]{
  \includegraphics[width=1.3in]{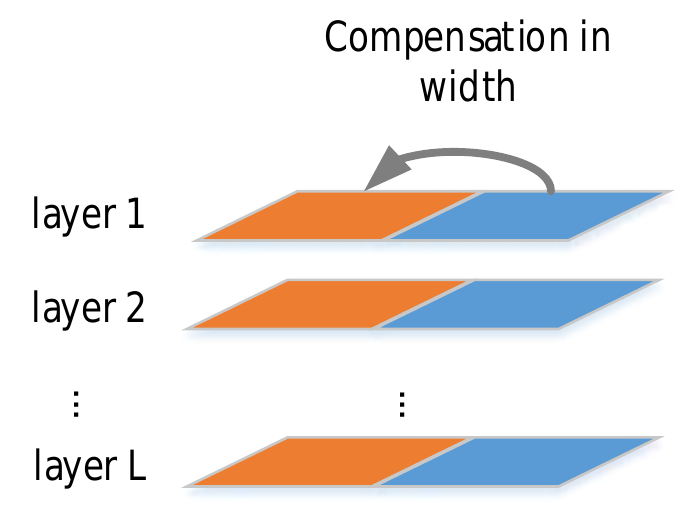}\label{fig.1.1}}
  \subfigure[Multi-level]{
  \includegraphics[width=1.22in]{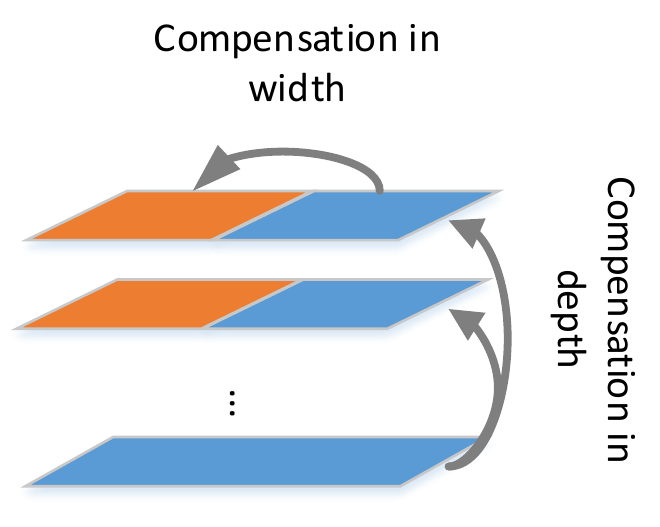}\label{fig.1.2}}
  \caption{Comparison of single-level quantization and multi-level quantization. The blue parts indicate the full-precision weights and layers of the network. The orange parts are quantized weights and layers.}\label{fig.1}
\end{figure}

One major challenge for network compression is the tradeoff between complexity and accuracy. However, most of the recent network compression methods degrade the accuracy of the network more or less \cite{han2015learning,guo2016dynamic,gong2014compressing,rastegari2016xnor,li2016ternary,zhou2016dorefa}. Recently, \cite{zhou2017incremental} propose incremental network quantization which re-trains the un-quantized parameters to compensate for the quantization loss can achieve high compression rate while maintaining performance. However, they pay no attention to the distribution of parameters and treat all layers equally (as shown in Figure \ref{fig.1.1}).

In this paper, we argue that both of the width level (partitioning parameters) and the depth level (partitioning layers) are important in network quantization (shown in Figure \ref{fig.1.2}). In the width level, quantization should fit the distribution of the weights which directly affects the accuracy of the network. Vector quantization is a quantization method that fully considers the distribution and makes quantization loss easy to be controlled. Furthermore, weights with special type may have special use ($e.g.$ weights with type powers of two may accelerate computations in FPGA devices). We extend our approach by using L1 norm to constrain the clustering process. In the depth level, layers are important elements of networks. They are interacted and make joint contributions to the networks. Thus, the quantization loss of one layer can be eliminated by re-training other layers. For ternary quantization, the huge quantization loss can not be compensated by re-training if only considering the width level. Thus, we introduce incremental layer compensation that quantize the layers partially and retrain other layers to compensate for the quantization loss. Considering both width level and depth level, the accuracy can be recovered after iteratively ternary quantization.

In summary, our contributions to network compression are two folds: (1) We propose single-level quantization approach for high-bit quantization. (2) For extremely low-bit quantization (ternary), we propose multi-level network quantization.

In the rest of the paper, we first introduce some related works and propose the single-level quantization approach. Next, we introduce the multi-level approach. Finally, we give the experiment results and the conclusion of the paper.

\section{Related Work}
\textbf{Compression by Low-rank Decomposition.} Reducing parameter dimensions using techniques like Singular Value Decomposition (SVD) \cite{denil2013predicting} works well on fully-connected layers and can achieve 3$\times$ compression rate. \cite{yu2017compressing} introduce this idea to convolutional layers by noting that weight filters usually share smooth components in a low-rank subspace and also remember some important information represented by weights that are sparsely scattered outside the low-rank subspace. Although this kind of method can achieve relatively good compression rate, the accuracy of some neural network models can be hurt.

\textbf{Compression by Pruning.} Pruning is a straightforward method to compress the networks by removing the unimportant parameters or convolutional filters. \cite{han2015learning} present an effective unstructured method to prune the parameters with values under a threshold and they reduce the model size by 9$\times$ on AlexNet and 16$\times$ on VGG-16. Structured pruning can greatly reduce the computation cost. \cite{li2016pruning} prune filters with small effect on the accuracy of the model and reduce the computation cost for VGG-16 by up to 34$\%$ and ResNet-110 by up to 38$\%$. \cite{luo2017thinet} further proposes a heuristic method to boost the filter level pruning.
In \cite{he2017channel}, channels are pruned and the pruning problem is formulated as a data recovery problem. The pruned models are accelerated by a large margin.

\textbf{Compression by Quantization.} Quantization is a many-to-few mapping between a large set of input and a smaller set of output. It groups weights with similar values to reduce the number of free parameters. Hash-net \cite{chen2015compressing} constrains weights hashed into different groups before training. Within each group the weights are shared and only the shared weights and hash indices need to be stored. \cite{gong2014compressing} compress the network with vector quantization techniques. \cite{han2015deep} present deep compression which combines the pruning \cite{han2015learning}, vector quantization and Huffman coding, and reduces the model size by 35$\times$ on AlexNet and 49$\times$ on VGG-16. However, these quantization methods takes time and will more or less hurt the performance of the network. Recently, \cite{zhou2017incremental} present incremental network quantization (INQ) method. This method partitions the weights into two different parts: one part is used to quantize and another part is used to retrain to compensate for quantization loss. The weights of the network are quantized incrementally and finally the accuracy of the quantized model is even higher the the original one. This method basically solves the problem of accuracy loss during network compression. However, in this paper, the values in the codebook are pre-determined and quantization group is handcrafted. Thus, this kind of quantization is not data based and the quantization loss can not be controlled. Besides, they only partition weights which we refer to the width level and can not achieve great result in extremely low-bit quantization.

\textbf{Compression by other strategies.} Some other people are trying to design DNNs with low precision weights, gradients and activations. \cite{rastegari2016xnor} propose Xnor-Net which is a network with binary weights and even binary inputs. \cite{tang2017train} discuss the basic elements of training a high accuracy binary network. \cite{li2016ternary} design ternary weight network. \cite{cai17hwgq} propose HWQN with low bit activations. Knowledge transfer is another method to train a small network. Knowledge Distilling \cite{hinton2015distilling} is proposed to distill the knowledge from an ensemble of models to a single model by imitate the soft output of them. Neuron selectivity transfer method \cite{huang2017like} explores a new kind of knowledge neuron selectivity to transfer the knowledge from the teacher model to the student model and achieves better performance. Specific DNN architectures are designed for mobile devices. \cite{howard2017mobilenets} propose MobileNets which apply depth-wise separable convolution to factorize a standard convolution into a depthwise convolution and a $1\times1$ convolution and show the effectiveness of such architecture across a wide range of applications. \cite{zhang2017shufflenet} present ShuffleNet. They apply group convolutions to pointwise convolutions and introduce shuffle operations to maintain the connections between groups which achieves $13\times$ speed up in ALexNet.

\section{Overview}
\begin{figure}[!htp]
  \centering
  \includegraphics[width=3in]{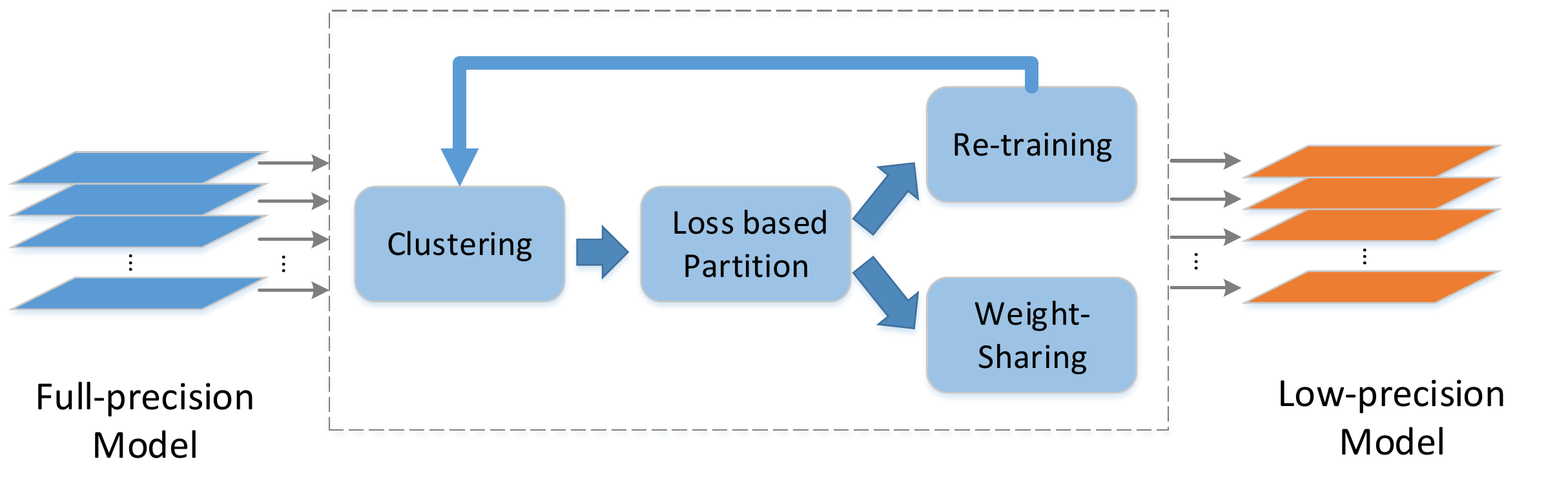}\\
  \caption{Framework of the proposed approach.}\label{fig.2}
\end{figure}

The framework of our approach is shown in Figure \ref{fig.2}. Either single-level quantization or multi-level quantization is composed of four steps: clustering, loss based partition, weight-sharing and re-training. Clustering uses k-means clustering to cluster the weights into $k$ clusters layer-wise. Loss based partition divides the $k$ clusters of each layer into two disjoint groups based on their quantization loss. The weights in one group are quantized into the centroids of their corresponding clusters by the weight-sharing step. The weights the other group are re-trained. Furthermore, all of the four steps are iteratively conducted until all the weights are quantized. The mainly difference for SLQ and MLQ is the loss based partition step. For SLQ, we only partition clusters. While for MLQ, we partition clusters and layers. Actually, SLQ is a particular case of MLQ. Technique details are discussed in the next sections.
\section{Single-level Quantization}

\subsection{Clustering}
\begin{figure}
  \centering
  \subfigure[]{
  \includegraphics[width=1.6in]{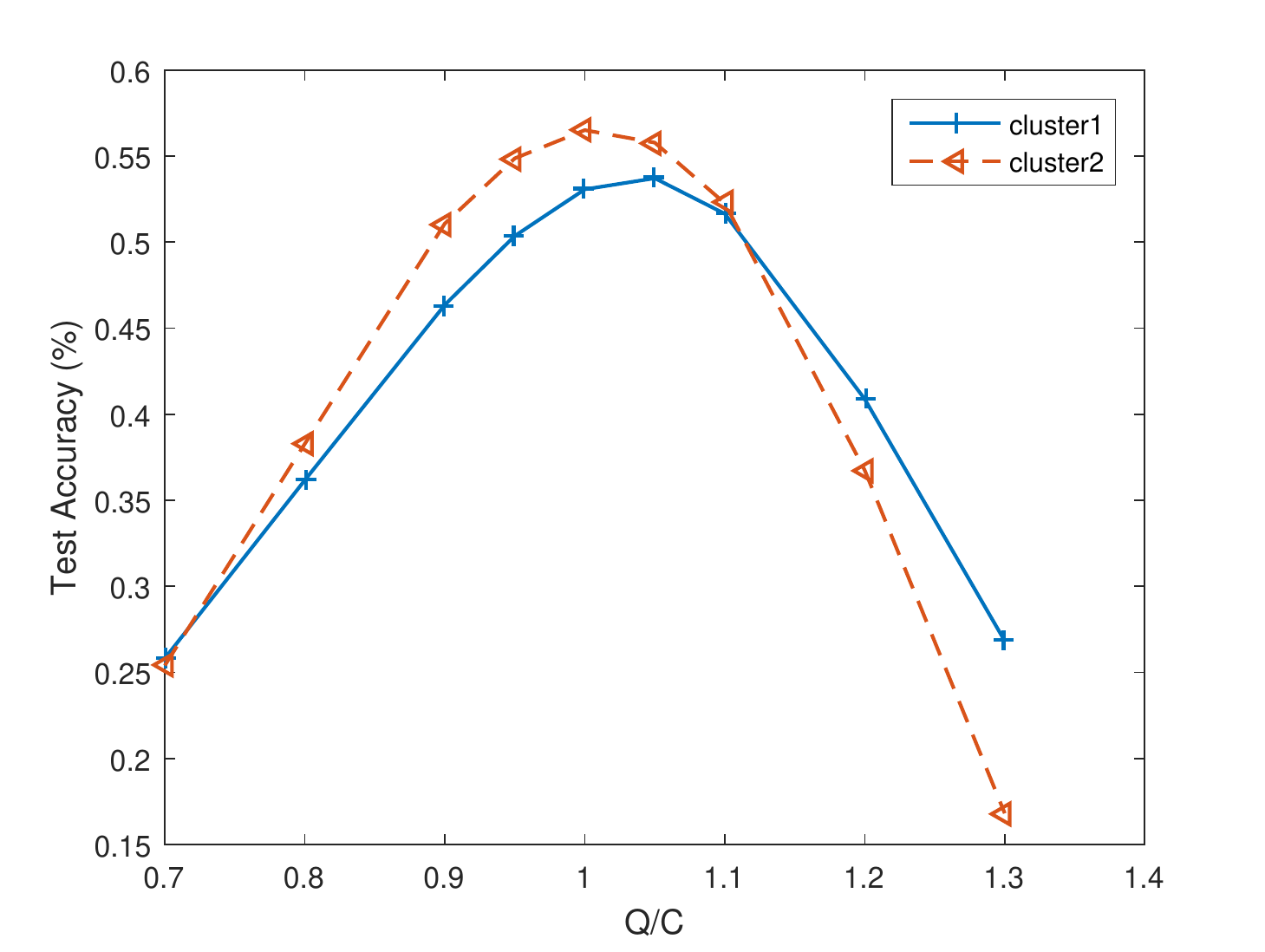}\label{fig.3.1}}
  \subfigure[]{
  \includegraphics[width=1.6in]{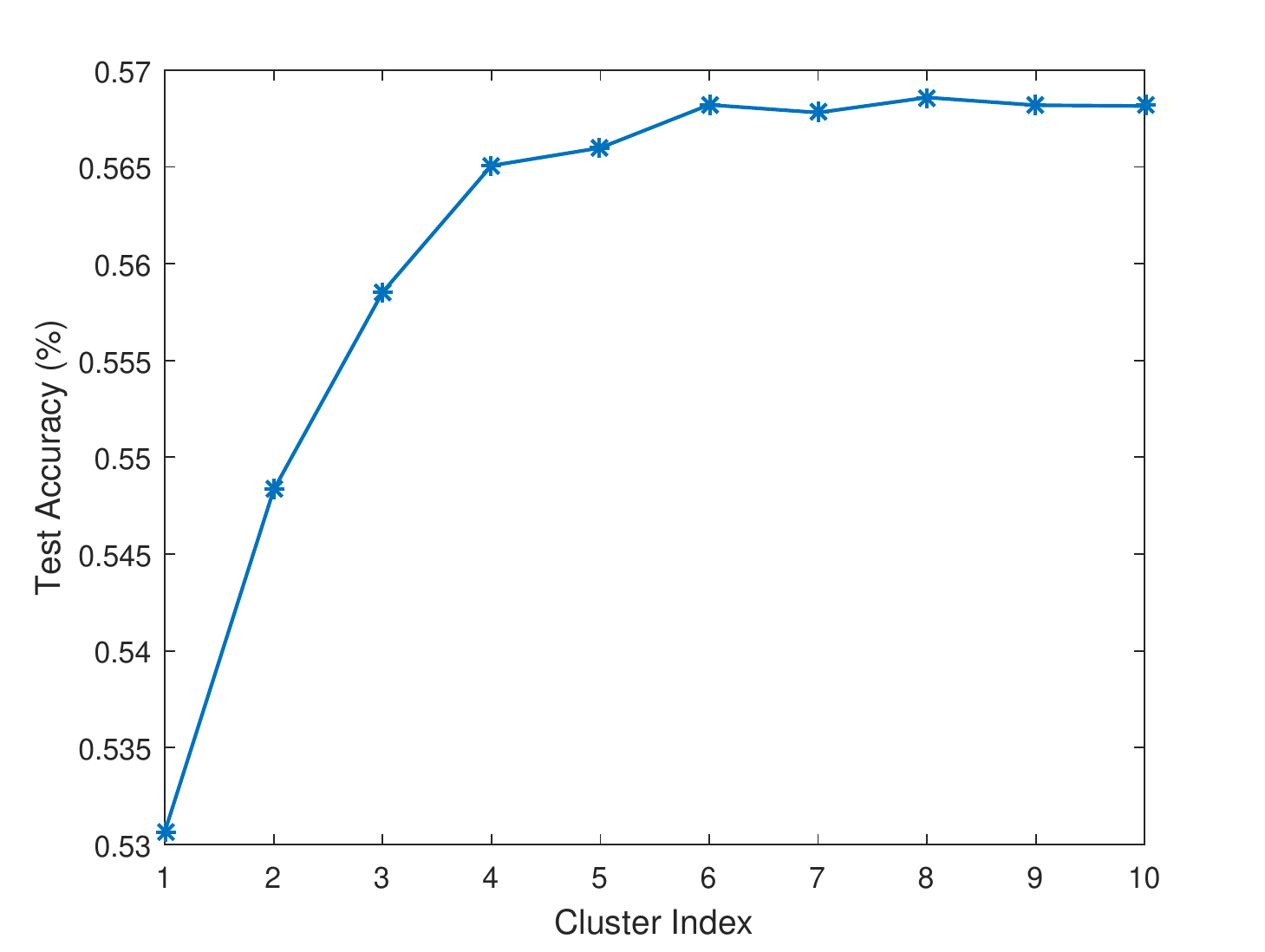}\label{fig.3.2}}
  \caption{(a) shows the quantization of two different clusters generated by k-means in AlexNet. Q/C means the value Q that weights to be quantized into is divided by the centroid of the cluster C.The accuracy of the network changes with the change of Q. (b) shows the test accuracy when 10 clusters of AlexNet are quantized respectively. The clusters are sorted in the descending order. }\label{fig.3}
\end{figure}

Other than using handcrafted mapping rules \cite{zhou2017incremental}, we adopt k-means clustering which is more data-driven and can easily control quantization loss. We choose two clusters generated by k-means and quantize the weights into different values including the centroids of them. Figure \ref{fig.3.1} shows that the quantization loss is low if we quantize the weights into the centroids of the clusters.

\begin{figure*}
  \centering
  \includegraphics[width=5in]{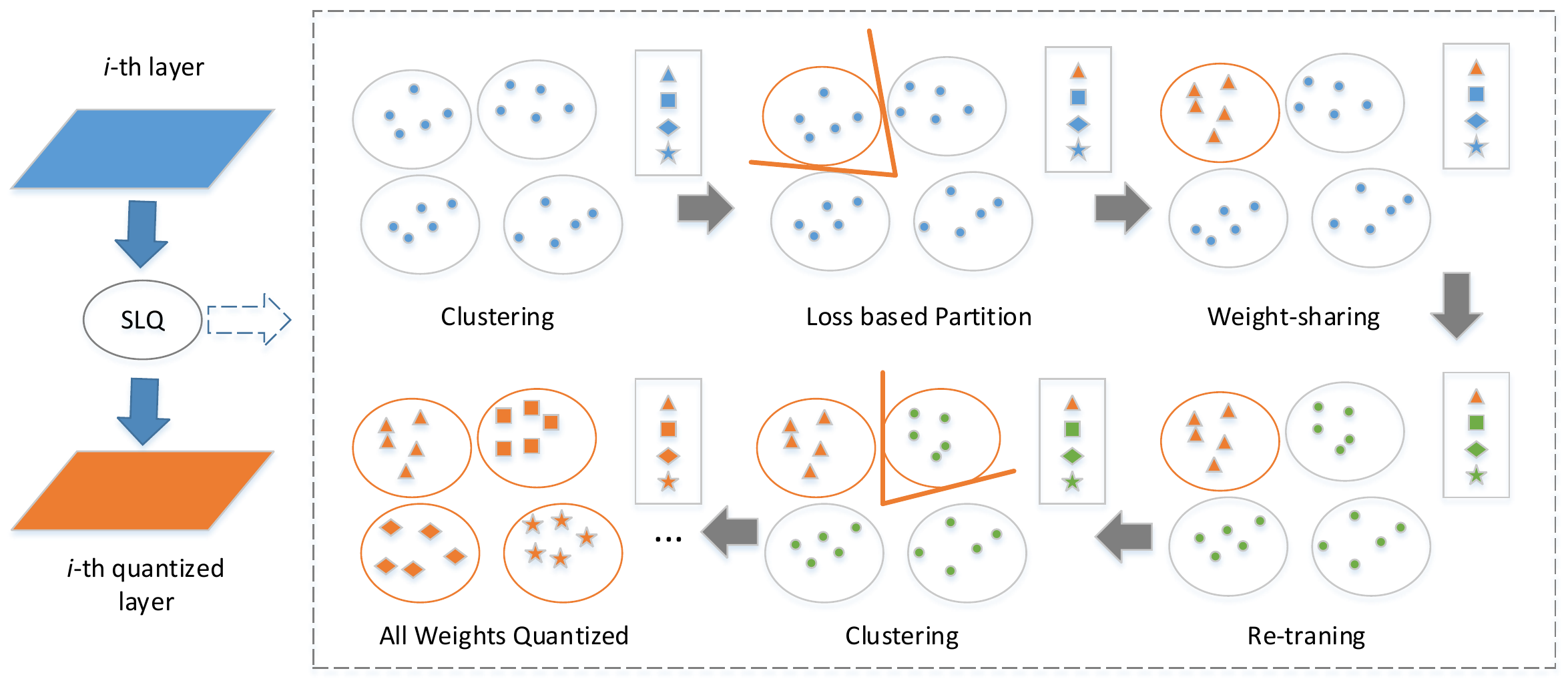}\\
  \caption{An schematic diagram of our single-level quantization approach: The small rectangle is the codebook. Blue, green and orange points indicates the full precision, re-trained and quantized weights. Clustering is conducted on pre-trained full-precision network; Performing loss based partition on the clusters; One group of clusters with weights are quantized into the centroids of clusters; Fixing quantized weights, the other clusters are re-trained; The re-trained weights are clustered by k-means clustering; After several iterations, all weights are quantized into the centroids. }\label{fig.4}
\end{figure*}

\subsection{Loss based Partition}
After the layer-wise clustering, each layer holds a code book, $\{c_{1}^{i},c_{2}^{i}\ldots c_{k}^{i}\},i=1,2\ldots L$, where $c_{k}^{i}$ denotes the $k^{th}$ centroid in the code book of $i^{th}$ layer. We partition the weights into two groups: the weights in one group are quantized and the weights in the other are re-trained. \cite{zhou2017incremental} use the pruning inspired strategy \cite{han2015learning} that weights with bigger values are more important and need to be quantized prior. However, this strategy is not suitable for our approach because the accuracy of the network can be affected by many factors during quantization including the value to be quantized into (as shown in Figure \ref{fig.3.1}) and the number of weights to be quantized. We test the quantization loss of 10 different clusters of AlexNet that generated by k-means. The result is shown in Figure \ref{fig.3.2}. There exist some clusters that do not fit the pruning inspired strategy \cite{zhou2017incremental}. Benefit from clustering, the weights are roughly partitioned and we only need to further partition the clusters. Besides, for the fact that the number of the clusters is relatively small, we propose loss based partition. We test the quantization loss of each cluster and sort the clusters by quantization loss. Cluster with bigger quantization loss is quantized prior.

For the $i^{th}$ layer, the loss based partition can be defined as:
\begin{equation}\label{equ.1}
\begin{split}
  &\mathbf \Phi_{(1)}^{i}\cup \mathbf \Phi_{(2)}^{i}=\mathbf W_{i},\quad \mathbf \Phi_{(1)}^{i}\cap \mathbf \Phi_{(2)}^{i}=0\\
  &s.t.\quad \min_{\mathbf\Phi_{(1)}^{i}} EQ>\max_{\mathbf\Phi_{(2)}^{i}} EQ
\end{split}
\end{equation}
where $\mathbf \Phi_{(1)}^{i}$ is the group containing the clusters to be quantized, while $\mathbf \Phi_{(2)}^{i}$ is the group containing the clusters to be re-trained. $\mathbf W_{i}$ is the set that covers all of the weights in the $i^{th}$ layer. $EQ$ is quantization loss of the cluster. The minimum $EQ$ of the clusters in $\mathbf \Phi_{(1)}^{i}$ is bigger than the maximum $EQ$ of clusters in $\mathbf \Phi_{(2)}^{i}$.

The clusters are partitioned into two groups, meanwhile the code book is also divided into two parts: one part is fixed while the other is updated.

\subsection{Weight-sharing}
We quantize the weights in the group $\mathbf \Phi_{(1)}^{i}$ by weight-sharing. The weights in this group are quantized into the centroids of the corresponding clusters. The weight-sharing of $i^{th}$ layer is described in Equation \ref{equ.2}.
\begin{equation}\label{equ.2}
\begin{split}
  &\omega(p,q)=c_{j}^{i},\\
  &s.t.\quad \omega(p,q)\in \mathbf \Psi_{j}^{i},\mathbf \Psi_{j}^{i}\in \mathbf\Phi_{(1)}^{i}
\end{split}
\end{equation}
where $\mathbf \Psi_{j}^{i}$ is the cluster in the quantization group $\mathbf\Phi_{(1)}^{i}$, while $c_{j}^{i}$ is the centroid of $\mathbf \Psi_{j}^{i}$.

\subsection{Re-training}

As weight-sharing brings error to the network, we need to re-train the model to recover accuracy. Thus, we fix the quantized weights and re-train the weights in the other group.
After re-training, as shown in Figure \ref{fig.4}, we will come back to beginning of our approach (clustering) to quantize the left weights iteratively until all the weights are quantized.

Taking the $l^{th}$ layer as an example, we use $Q_{l}$ to denote the set of quantized weights in the $l^{th}$ layer. To simplify the problem, we define a mask matrix $M_{l}(p,q)$, which has the same size as weight matrix $\omega_{l}(p,q)$ and acts  as an indicator function to indicate that if the weights has been quantized. $M_{l}(p,q)$ can be defined as:
\begin{equation}\label{equ.3}
M_{l}(p,q)=\left\{
\begin{aligned}
0 & \quad , if\quad \omega_{l}(p,q)\in Q_{l}\\
1 &  \quad , otherwise
\end{aligned}
\right.
\end{equation}

During re-training, our quantization approach can also be treated as an optimization problem:
\begin{equation}\label{equ.4}
\begin{split}
  \min_{\omega_{l}}\quad &E(\omega_{l})=L(\omega_{l})+\sum_m \lambda_m R_m(\omega_{l})\\
  s.t.\quad &\omega_{l}(p,q)\in B_{l},if\ M_{l}(p,q)=0
\end{split}
\end{equation}

where $L(\omega_{l})$ is the loss of the network, $R_m(\omega_{l})$ is the regulation term of the $m^{th}$ iteration that constrains the weights to be quantized into the centroids within $B_{lm}$. $\lambda_m$ is a positive scalar. ${B_{l}}$ is the codebook of the centroids after $m$ iterations.

To solve the optimization problem, we re-train the network using stochastic gradient decent(SGD) to update the un-quantized weights. To fix the quantized weights, we use the indicator function $M_{l}(p,q)$ as a mask on the gradient of the weights to control the gradient propagation:
\begin{equation}\label{equ.5}
\begin{split}
  \omega_{l}(p,q)\leftarrow \omega_{l}(p,q)-\gamma \frac{\partial E}{\partial (\omega_{l}(p,q))}M_{l}(p,q)
\end{split}
\end{equation}

The whole quantization process is shown in Algorithm \ref{alg1}.

\subsection{Extended Approach}

Later, we extend single-level quantization (SLQ) approach. In the SLQ approach, we quantize the weights layer-wise into the centroids of clusters. However, sometimes we need the weights to be some special type. For instance, if all the weights are power of two, the model will be convenient to be deployed in FPGA devices.

The main difference of our extended single-level quantization (ESLQ) with original SLQ is that we extend traditional clustering to constrain the cluster centroid to close or equal to the number with oriented type ($t$-centroid). Thus, after weight-sharing, we can quantize the weights into values with oriented type. For instance, we want to constrain centroid $c_{1}^{i}$ to close to or equal to a specific type: $t$-centroid $\hat{c}_{1}^{i}$. We incorporate the L1 norm regulation into the traditional k-means loss function as:
\begin{equation}\label{equ.6}
\begin{split}
\min_{\mathbf \Psi_{1}^{i},\mathbf \Psi_{2}^{i}\ldots \mathbf \Psi_{k}^{i}}&\frac{1}{|\omega^{i}|}\sum_{j=1}^{k}\sum_{\omega(p,q)\in \mathbf \Psi_{j}^{i}}|\omega(p,q)-c_{j}^{i}|^{2}+\beta_{1}|c_{1}^{i}-\hat{c}_{1}^{i}|, \\
   s.t.\quad & c_{j}^{i}=\frac{1}{|\mathbf \Psi_{j}^{i}|}\sum_{\omega(p,q)\in \mathbf \Psi_{j}^{i}}\omega(p,q), i=1,2\ldots L
\end{split}
\end{equation}
where $\hat{c}_{1}^{i}$ is the $t$-centroid of $c_{1}^{i}$, $|\omega^{i}|$ denotes the total number of weights in the $i^{th}$ layer. We weighted the original k-means loss by $\frac{1}{|\omega^{i}|}$ to strengthen the impact of the regularization term.

In ESLQ, we first conduct traditional clustering and loss based partition. Then we determine the $t$-centroids of the cluster to be quantized. Subsequently, we re-cluster the weights by our extended clustering. The weight-sharing and re-training steps are the same as SLQ. After several iterations, the network can be quantized into oriented type.

{
\begin{algorithm}[t]
\renewcommand{\baselinestretch}{1.0}
\renewcommand{\arraystretch}{1.0}
\small
\caption{Single-Level Quantization}\label{alg1}
\begin{algorithmic}[1]
\STATE\textbf{Input:}  $\left\{\omega_{l}:1\leq l\leq L\right\}$: the pre-trained full-precision DNN model\\
\STATE\textbf{output:}  $\left\{\omega_{l}^{'}:1\leq l\leq L\right\}$: the final low-precision model with the weights quantized into the centroids in code book ${B_{l}}$\\
\FOR {$m=1,2,\dots,N$}
\STATE Reset the base learning rate and the learning policy
\STATE Apply k-means clustering layer-wise
\STATE Perform loss based partition layer-wise by Equation \ref{equ.1}
\STATE Quantize the weights in one group by Equation \ref{equ.2}
\STATE Re-train the network as described in the Re-training section
\ENDFOR
\end{algorithmic}
\end{algorithm}
}
\section{Multi-Level Quantization}

The proposed SLQ approach is not suitable for low-bit quantization ($e.g.$ 2-bit quantization into ternary networks) because the number of clusters is small and the quantization loss in each iteration step is too huge to be eliminated. We introduce incremental layer compensation (ILC) to partition the layers of the network which is the depth level of the network. The ILC is motivated by the intuition that different layers have different impact on the performance of the network during quantization, $e.g.$ convolutional layers and fully connected layers. The layers $L$ of the network are partitioned into two groups: one group $L_q$ containing layers with more quantization loss is quantized prior and another group $L_r$ containing the remaining layers is re-trained:

\begin{equation}\label{equ.7}
  L_q \cup L_r = L, and \quad L_q \cap L_r=0\\
\end{equation}

\begin{figure}
  \centering
  \includegraphics[width=3in]{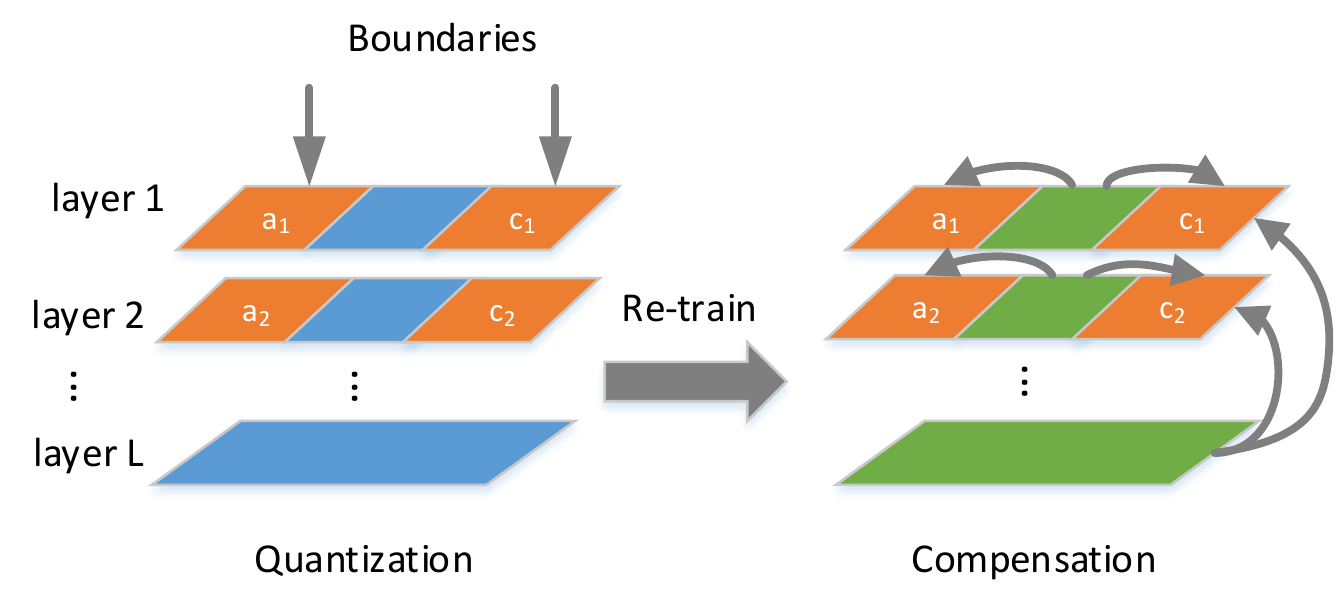}\\
  \caption{Quantization process of multi-level ternary quantization. Blue, green and orange parts indicates the full precision, re-trained and quantized layers. We first quantize the Boundaries and then quantize the Hearts of the network.}\label{fig.5}
\end{figure}

We introduce ILC into SLQ which is multi-level quantization (shown in Figure \ref{fig.5}). The MLQ partitions both the layers and the parameters within layers, which lowers the huge quantization loss in low-bit quantization ($e.g.$ 2-bit quantization). Taking the $i^{th}$ layer as an example (ternary quantization), each layer is clustered into 3 clusters and we obtain three centroids: $a_i, b_i$ and $c_i$. $a_i$ and $c_i$ affect the performance of the networks more. We call them Boundaries. $b_i$ holding smaller effect is called Heart. We first quantize Boundaries of the network. Different from SLQ that quantizes all the Boundaries at the same time, the MLQ quantizes the boundaries iteratively by ILC. The Boundaries in different layers are partitioned into two groups, one group is quantized and the remaining weights in the network are all re-trained. After all the boundaries are quantized, we then quantize the Hearts iteratively by ILC too. After several iterations, the Boundaries and the Hearts are all quantized (shown in Algorithm \ref{alg2}).

{
\begin{algorithm}[t]
\renewcommand{\baselinestretch}{1.0}
\renewcommand{\arraystretch}{1.0}
\small
\caption{Multi-Level Quantization}
\label{alg2}
\begin{algorithmic}[1]
\STATE\textbf{Input:} $\left\{\omega_{l}:1\leq l\leq L\right\}$: the pre-trained full-precision CNN model
\STATE\textbf{output:} $\left\{\omega_{l}^{'}:1\leq l\leq L\right\}$: the ternary network
\STATE Apply k-means clustering layer-wise (cluster number is 3)
\STATE Perform loss based partition layer-wise to generate Boundaries and Hearts of the network
\STATE Quantize the Boundaries iteratively by ILC (partition, weight-sharing and re-training)
\STATE Quantize the Hearts iteratively by ILC (partition, weight-sharing and re-training)
\end{algorithmic}
\end{algorithm}
}

\section{Experiments}
To analyze the performance of SLQ and MLQ, we conduct extensive experiments on two datasets: CIFAR-10 and ImageNert.

The bit-width parameter b represents the space we used to store each quantized weight. To fairly compared with other methods, we use b bits to code the centroids: one bit to store zero and the other (b-1) bits to code non-zero centroids which means that for bit-width b, the centroid number of each layer is $2^{b-1}+1$.

\textbf{CIFAR-10}: This dataset consists of 60,000 32$\times$32 colour images in 10 classes, with 6000 images per class. There are 50,000 training images and 10,000 test images.

\textbf{ImageNet}: This dataset contains as much as 1000 classes of objects with nearly 1.2 million training images and 50 thousand validation images.

\begin{figure}
  \centering
  \subfigure[Light CNN training]{
  \includegraphics[width=1.6in]{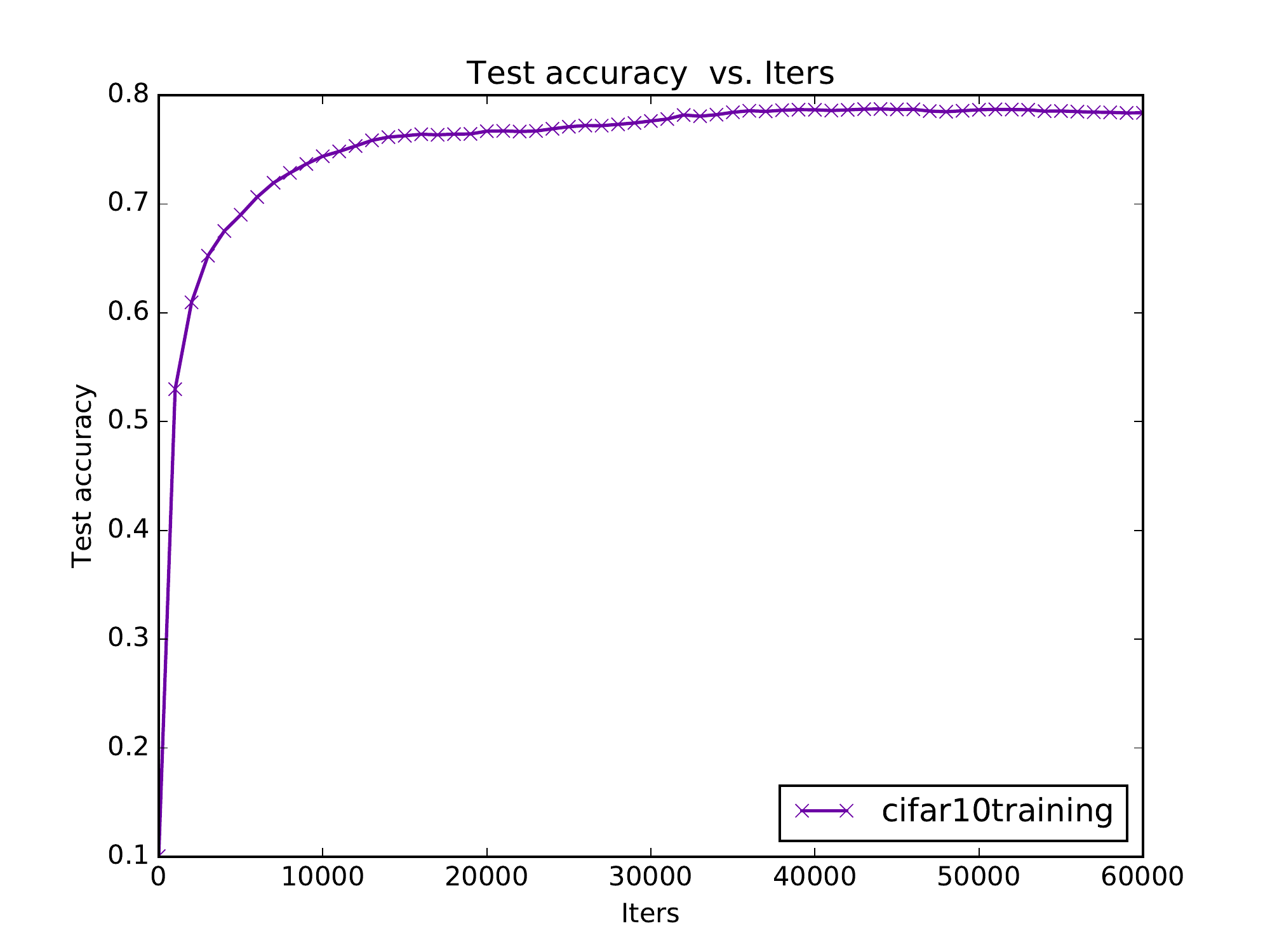}\label{fig.6.1}}
  \subfigure[Light CNN quantization]{
  \includegraphics[width=1.6in]{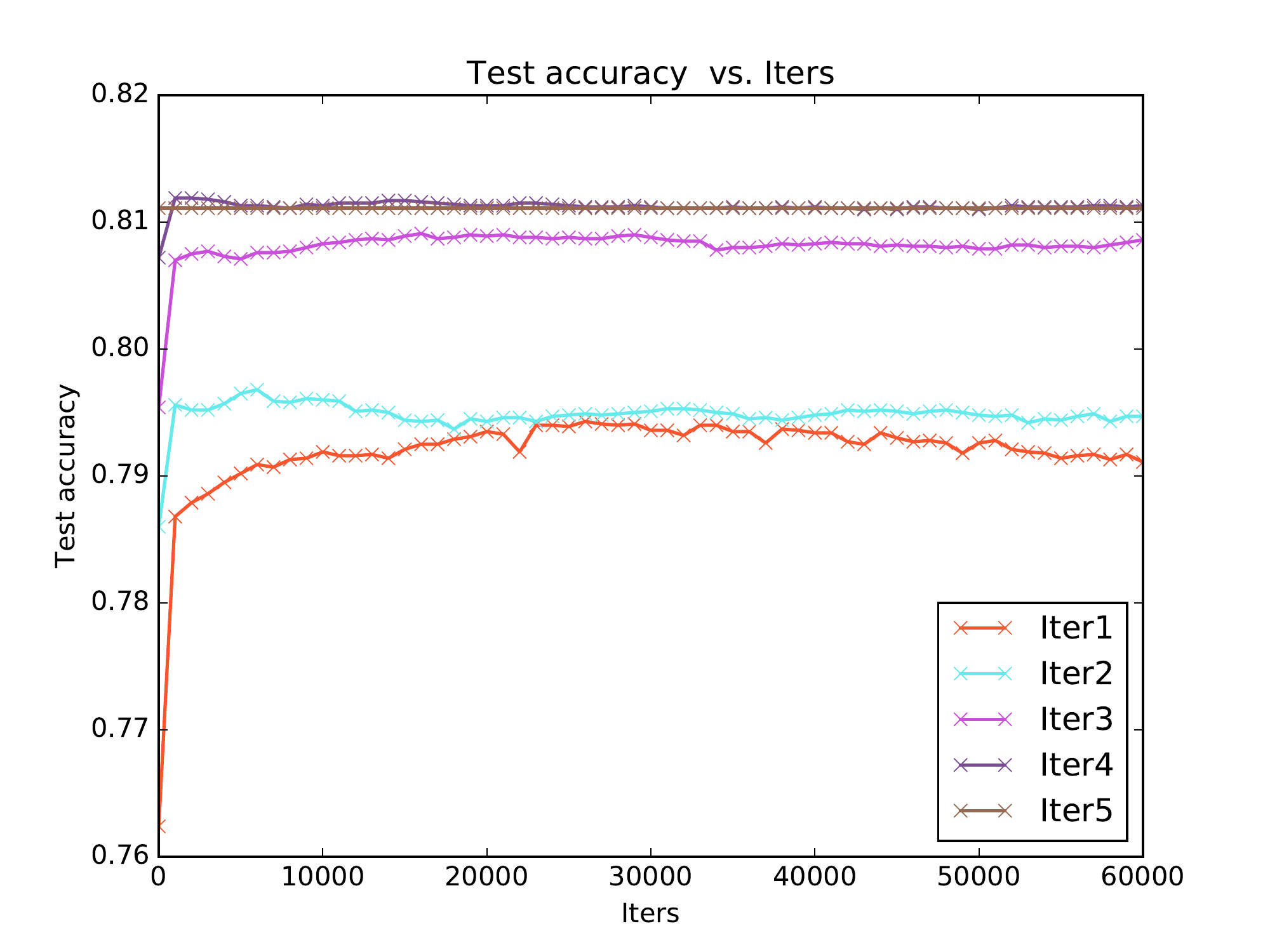}\label{fig.6.2}}
  \caption{(a) is the training curves of the light CNN. (b) is the training curves in 5 iterations of SLQ quantization on light CNN.}\label{fig.6}
\end{figure}

\begin{table}[!htp]
  \centering

  \begin{tabular}{c c c c }

\hline
 Network  &Bit-width     & Accuracy &Increase \\
\hline
Light CNN ref  &32& 78.66$\%$&\\
Light CNN SLQ&5&\textbf{81.11$\%$}&2.45$\%$\\
\hline
\hline
ResNet20 ref&32&91.70$\%$&\\
ResNet20 SLQ&5&\textbf{91.75$\%$}&0.05$\%$\\
\hline
\end{tabular}
\caption{Experiment results of 5-bit SLQ on CIFAR-10. }\label{table.1}
\end{table}

\begin{table*}[!htp]
  \centering

  \begin{tabular}{c c c c c c c}

\hline
 Network  &Batch size &Weight decay    & momentum&Bit-width&Cluster number&Partition\\
\hline
AlexNet  &256&0.0005&0.9&5&17&5,4,4,2,2\\
VGG16&32&0.0005&0.9&5&17&5,4,4,2,2\\
GoogleNet&80&0.0005&0.9&5&17&5,4,4,2,2\\
ResNet18&80&0.0005&0.9&5&17&5,4,4,2,2\\
\hline

\end{tabular}
\caption{Parameter settings of networks. }\label{table.2}
\end{table*}

\subsection{Results for SLQ}

\begin{table*}[!htp]
  \centering

  \begin{tabular}{c c c c c c}

\hline
Network  &Bit-width&Cluster number     & Top1 accuracy &Top5 accuracy&Increase in top-1/top-5 error\\

AlexNet ref  &32&& 57.10$\%$&80.20$\%$\\

AlexNet INQ&5&17&57.39$\%$ &80.46$\%$\\

ALexNet SLQ&5&17&\textbf{57.56$\%$}&\textbf{80.50$\%$} &0.46$\%$/0.30$\%$\\
\hline
\hline
VGG16 ref  &32&& 68.54$\%$&88.65$\%$\\

VGG16 INQ&5&17&70.82$\%$ &90.30$\%$\\

VGG16 SLQ&5&17&\textbf{72.23$\%$}&\textbf{91.0$\%$} &3.69$\%$/1.35$\%$\\
\hline
\hline
Googlenet ref  &32&& 68.89$\%$&89.03$\%$\\

Googlenet INQ&5&17&69.02$\%$ &89.28$\%$\\

Googlenet SLQ&5&17&\textbf{69.10$\%$}&89.19$\%$ &0.21$\%$/0.16$\%$\\
\hline
\hline
Resinet18 ref  &32&& 68.27$\%$&88.69$\%$\\

Resinet18 INQ&5&17&68.98$\%$ &89.10$\%$\\

Resinet18 SLQ&5&17&\textbf{69.09$\%$}&\textbf{89.15$\%$} &0.82$\%$/0.46$\%$\\
\hline
\end{tabular}
\caption{Experiment results of SLQ method on ImageNet. }\label{table.3}
\end{table*}

\begin{table*}[!htp]
  \centering

  \begin{tabular}{c c c c c c}

\hline
 Network  &Bit-width &Centroid number    & Top-1 accuracy &Top-5 accuracy&Increase in top-1/top-5 error\\
\hline
VGG16 ref  &32&& 68.54$\%$&88.65$\%$\\
VGG16 SLQ&5&17&72.23$\%$&91.0$\%$ &3.69$\%$/1.35$\%$\\
VGG16 SLQ&4&9&71.18$\%$&90.25$\%$&2.64$\%$/0.60$\%$\\
VGG16 SLQ&3&5&68.38$\%$&88.55$\%$&-0.16$\%$/-1.10$\%$\\
\hline
\end{tabular}
\caption{Experiment results of bit-width change on ImageNet. }\label{table.4}
\end{table*}

\begin{table*}[!htp]
  \centering

  \begin{tabular}{c c c c c c}

\hline
 Network  &Bit-width &Cluster number    & Top-1 accuracy &Top-5 accuracy&Increase in top-1/top-5 error\\
\hline
VGG16 ref  &32&& 68.54$\%$&88.65$\%$\\
VGG16 non-linear&5&17&72.23$\%$&91.0$\%$ &3.69$\%$/1.35$\%$\\
VGG16 linear&5&17&71.85$\%$&90.87$\%$&3.31$\%$/1.22$\%$\\
\hline
\end{tabular}
\caption{Experiment results of centroid initialization of SLQ. }\label{table.5}
\end{table*}

\begin{table*}[!htp]
  \centering

  \begin{tabular}{c c c c c c}

\hline
 Network  &Bit-width&Cluster number     & Top-1 accuracy &Top-5 accuracy&Increase in top-1/top-5 error\\
\hline
AlexNet ref  &32&& 57.10$\%$&80.20$\%$\\

AlexNet ESLQ1  &5&17&57.26$\%$&80.28$\%$&0.16$\%$/0.08$\%$\\
AlexNet ESLQ2  &5&17&57.42$\%$&80.25$\%$&0.32$\%$/0.05$\%$\\
\hline
\hline
VGG16 ref  &32&& 68.54$\%$&88.65$\%$\\
VGG16 ESLQ1 &5&17&71.17$\%$&90.50$\%$&2.63$\%$/0.85$\%$\\
VGG16 ESLQ2 &5&17&71.95$\%$&90.86$\%$&3.41$\%$/1.21$\%$\\
\hline
\end{tabular}
\caption{Experiment results of ESLQ method on ImageNet. }\label{table.6}
\end{table*}

\subsubsection{SLQ Results on CIFAR-10}
We use the light CNN (three convolutional layers and three fully connected layers) offered in Caffe \cite{jia2014caffe} and ResNet20 \cite{he2016deep} to conduct the classification on CIFAR-10. The light CNN is trained from scratch (as shown in Figure \ref{fig.6.1}). After 5 iterations the trained full-precision light CNN model is quantized into 5-bit low-precision model (shown in Figure \ref{fig.6.2}). The quantization loss of each iteration is decreasing. The quantization results of two networks are shown in Table \ref{table.1}. Both of the two networks enjoy accuracy increase after quantization by SLQ.

\subsubsection{SLQ Results on ImageNet}
We apply the proposed SLQ approach to various popular models on ImageNet including: AlexNet \cite{Krizhevsky2012ImageNet}, VGG-16 \cite{simonyan2014very}, GoogleNet \cite{szegedy2015going} and Resinet-18 \cite{he2016deep}. All these full-precision networks are quantized into 5-bit low precision ones. The setting of the parameters is shown in Table \ref{table.2}. The cluster partition ways of the four networks are the same which means that our approach is easier to implement and is robust on different DNN architectures. The results are shown in Table \ref{table.3}. The 5-bit CNN models quantized by SLQ have better performance in the ImageNet large scale classification task both in Top1 and Top5 accuracy than full-precision references. We also compare our SLQ results with INQ \cite{zhou2017incremental}. Our approach achieves improvement in all of the Top1 accuracy and most of the Top5 accuracy. It shows that considering the distribution of weights during quantization is very important and the loss based partition also contributes to the increase.

\subsection{Results for SLQ with Low-bit Setting}

In this experiment, we test our SLQ approach in different bit-width settings. We use VGG-16 as our test model. Except for the original 5-bit quantization result, we present 4-bit and 3-bit results which is shown in Table \ref{table.4}. As 5-bit compressed model, our 4-bit compressed model can also have good performance in both Top-1 and Top-5. However, for bit-width as low as 3 which means that the centroid number is 5, the accuracy of the compressed model drops a little. The loss based partition step in SLQ is related to the number of centroids. If the centroid number is big enough(for instance 17 and 9), we can have more iterations during quantization. While if the centroid number is small(for instance 5), we will have less iterative quantization steps and the quantization loss in the last quantization step is big. That is why the accuracy of the 3-bit compressed model is slightly lower than reference full-precision VGG-16 model. Thus, we have to try other ways ($e.g.$ our proposed MLQ) to conduct extremely low-bit quantization. The partition ways of the experiments are described bellow:

5-bit VGG-16 cluster partitions are $\{5, 4, 4, 2, 2\}$;

4-bit VGG-16 cluster partitions are $\{3, 2, 2, 2\}$;

3-bit VGG-16 cluster partitions are $\{2, 2, 1\}$.

\subsection{Results for Centroid Initialization}
We conduct experiments to show the effect of centroid initialization on our SLQ approach. We choose two kinds of centroid initialization ways. One is linear (linear decaying) and the other is non-linear (exponential decaying).

We choose VGG-16 as our test model. The results are shown in Table \ref{table.5}. The accuracy of the model quantized by SLQ with non-linear initialization is higher than the accuracy of SLQ with linear initialization. The centroid of the clusters to be quantized in the last iteration is smaller, so the number of weights is also smaller. This leads to the smaller quantization loss in the last iteration. Thus, we adopt non-linear initialization in all of our experiments.

\subsection{Results for ESLQ}

In this experiment, we test our ESLQ approach. The highlights of our ESLQ approach is to quantize the weights to oriented type: t-centroids. To test it, we choose two types: one is scientific notation with two significant figures and the other is either power of two or zero. The experiment results are shown in Table \ref{table.6}. In Table \ref{table.6}, ESLQ1 indicates the scientific notation and ESLQ2 indicates the power of 2.  The model quantized with ESLQ in both of the two situations have accuracy increase which shows the effectiveness of ESLQ.
\subsection{Results for MLQ}
We quantize light CNN and ResNet20 into ternary networks on CIFAR-10. In our experiments, we train the networks on CIFAR-10 without using data augmentation. The results are shown in \ref{table.7}. The accuracy of the ternary light CNN and ternary ResNet20 decrease little compared with the full-precision ones.

AlexNet model is quantized into ternary on ImageNet by MLQ. We compare our MLQ approach with TWN \cite{li2016ternary} and TTQ \cite{zhu2016trained} (shown in Table \ref{table.8}). Both TWN and TTQ add bach normalization layers by which the baseline of AlexNet can reach up to 60$\%$. Moreover, the batch normalization layers also contribute to the convergency of their network during training. In TTQ, they do not quantize the first convolutional layer and the last fully connected layer, that is another reason of their high performance. Different from them, our MLQ approach is more robust, we do not change the architecture of the network (without adding batch normalization layer) and quantize all of layers in ALexNet which can still achieve comparable results. Another method FGQ \cite{mellempudi2017ternary} conducts ternary quantization without additional training. Our method outperforms FGQ, though we have more training time cost.

\begin{table}[!htp]
  \centering

  \begin{tabular}{c c c c }

\hline
 Network  &Bit-width     & Accuracy &Increase \\
\hline
Light CNN ref  &32& 78.66$\%$&\\
Light CNN MLQ&2(ternary)&78.46$\%$&-0.2$\%$\\
\hline
\hline
ResNet20 ref&32&91.70$\%$&\\
ResNet20 MLQ&2(ternary)&90.02&-1.68$\%$\\
\hline
\end{tabular}
\caption{Experiment results of MLQ on CIFAR-10. }\label{table.7}
\end{table}

\begin{table}[!htp]
  \centering

  \begin{tabular}{c c c c c }

\hline
 Network& BN & Top-1    &Top-5  &Layers \\
\hline
Baseline&No&57.1$\%$&80.2$\%$&\\
TWN  &Yes&54.5$\%$&76.8$\%$ &8 layers\\
 TTQ &Yes&57.5$\%$&79.7$\%$ &6 layers\\
MLQ(ours)&No&54.24$\%$&77.78$\%$&8 layers\\
\hline
\end{tabular}
\caption{Experiment results of MLQ on ImageNet. }\label{table.8}
\end{table}

\subsection{Compression Ratio and Acceleration}
The compression ratio can be easily computed by the bit-width of the networks. The compression ratio of the 5-bit compressed AlexNet is 6$\times$. Besides, the proposed approach can be combined with the pruning strategy \cite{han2015learning} to further compress the network. The 5-bit pruned AlexNet is 53$\times$ compressed without accuracy loss. Since current BLAS libraries on CPU and GPU do not support indirect look-up and relative indexing, accelerators designed for quantized models \cite{han2016eie} can be adopted.

For training time, with one NVIDIA TITAN Xp, the proposed approach takes about 28 hours to accomplish 5-bit AlexNet quantization on ImageNet.

\section{Conclusion}
In this paper, we propose single-level quantization (SLQ) and multi-level quantization (MLQ) by considering the network quantization both from width and depth. By taking the distribution of the parameters into account, the SLQ obtains accuracy gain in the high-bit quantization of state-of-the-art networks on two datasets. Besides, the MLQ achieves impressive results in extremely low-bit quantization (ternary) without changing the architecture of networks.

\section{Acknowledgements}
This work is supported in part by NSFC (61425011, 61529101, 61720106001, 61622112, 61472234), the Program of Shanghai Academic Research Leader (17XD1401900) and Tencent research grant. We would like to thank Haoyang Yu and Xin Liu from Shenzhen Tencent Computer System Co.,Ltd. for their valuable discussions about the paper.
\bibliographystyle{aaai}
\bibliography{dnncompress}

\begin{thebibliography}{}

\bibitem[\protect\citeauthoryear{Cai \bgroup et al\mbox.\egroup
  }{2017}]{cai17hwgq}
Cai, Z.; He, X.; Sun, J.; and Vasconcelos, N.
\newblock 2017.
\newblock Deep learning with low precision by half-wave gaussian quantization.
\newblock In {\em CVPR}.

\bibitem[\protect\citeauthoryear{Chen \bgroup et al\mbox.\egroup
  }{2014}]{chen2014semantic}
Chen, L.-C.; Papandreou, G.; Kokkinos, I.; Murphy, K.; and Yuille, A.~L.
\newblock 2014.
\newblock Semantic image segmentation with deep convolutional nets and fully
  connected crfs.
\newblock {\em arXiv preprint arXiv:1412.7062}.

\bibitem[\protect\citeauthoryear{Chen \bgroup et al\mbox.\egroup
  }{2015}]{chen2015compressing}
Chen, W.; Wilson, J.; Tyree, S.; Weinberger, K.; and Chen, Y.
\newblock 2015.
\newblock Compressing neural networks with the hashing trick.
\newblock In {\em ICML},  2285--2294.

\bibitem[\protect\citeauthoryear{Denil \bgroup et al\mbox.\egroup
  }{2013}]{denil2013predicting}
Denil, M.; Shakibi, B.; Dinh, L.; de~Freitas, N.; et~al.
\newblock 2013.
\newblock Predicting parameters in deep learning.
\newblock In {\em NIPS},  2148--2156.

\bibitem[\protect\citeauthoryear{Girshick}{2015}]{Girshick_2015_ICCV}
Girshick, R.
\newblock 2015.
\newblock Fast r-cnn.
\newblock In {\em ICCV}.

\bibitem[\protect\citeauthoryear{Gong \bgroup et al\mbox.\egroup
  }{2014}]{gong2014compressing}
Gong, Y.; Liu, L.; Yang, M.; and Bourdev, L.
\newblock 2014.
\newblock Compressing deep convolutional networks using vector quantization.
\newblock {\em arXiv preprint arXiv:1412.6115}.

\bibitem[\protect\citeauthoryear{Guo, Yao, and Chen}{2016}]{guo2016dynamic}
Guo, Y.; Yao, A.; and Chen, Y.
\newblock 2016.
\newblock Dynamic network surgery for efficient dnns.
\newblock In {\em NIPS},  1379--1387.

\bibitem[\protect\citeauthoryear{Han \bgroup et al\mbox.\egroup
  }{2015}]{han2015learning}
Han, S.; Pool, J.; Tran, J.; and Dally, W.
\newblock 2015.
\newblock Learning both weights and connections for efficient neural network.
\newblock In {\em NIPS},  1135--1143.

\bibitem[\protect\citeauthoryear{Han \bgroup et al\mbox.\egroup
  }{2016}]{han2016eie}
Han, S.; Liu, X.; Mao, H.; Pu, J.; Pedram, A.; Horowitz, M.~A.; and Dally,
  W.~J.
\newblock 2016.
\newblock Eie: efficient inference engine on compressed deep neural network.
\newblock In {\em Proceedings of the 43rd International Symposium on Computer
  Architecture},  243--254.
\newblock IEEE Press.

\bibitem[\protect\citeauthoryear{Han, Mao, and Dally}{2015}]{han2015deep}
Han, S.; Mao, H.; and Dally, W.~J.
\newblock 2015.
\newblock Deep compression: Compressing deep neural networks with pruning,
  trained quantization and huffman coding.
\newblock {\em arXiv preprint arXiv:1510.00149}.

\bibitem[\protect\citeauthoryear{He \bgroup et al\mbox.\egroup
  }{2016}]{he2016deep}
He, K.; Zhang, X.; Ren, S.; and Sun, J.
\newblock 2016.
\newblock Deep residual learning for image recognition.
\newblock In {\em CVPR},  770--778.

\bibitem[\protect\citeauthoryear{He, Zhang, and Sun}{2017}]{he2017channel}
He, Y.; Zhang, X.; and Sun, J.
\newblock 2017.
\newblock Channel pruning for accelerating very deep neural networks.
\newblock In {\em ICCV}.

\bibitem[\protect\citeauthoryear{Hinton, Vinyals, and
  Dean}{2015}]{hinton2015distilling}
Hinton, G.; Vinyals, O.; and Dean, J.
\newblock 2015.
\newblock Distilling the knowledge in a neural network.
\newblock {\em arXiv preprint arXiv:1503.02531}.

\bibitem[\protect\citeauthoryear{Howard \bgroup et al\mbox.\egroup
  }{2017}]{howard2017mobilenets}
Howard, A.~G.; Zhu, M.; Chen, B.; Kalenichenko, D.; Wang, W.; Weyand, T.;
  Andreetto, M.; and Adam, H.
\newblock 2017.
\newblock Mobilenets: Efficient convolutional neural networks for mobile vision
  applications.
\newblock {\em arXiv preprint arXiv:1704.04861}.

\bibitem[\protect\citeauthoryear{Huang and Wang}{2017}]{huang2017like}
Huang, Z., and Wang, N.
\newblock 2017.
\newblock Like what you like: Knowledge distill via neuron selectivity
  transfer.
\newblock {\em arXiv preprint arXiv:1707.01219}.

\bibitem[\protect\citeauthoryear{Jia \bgroup et al\mbox.\egroup
  }{2014}]{jia2014caffe}
Jia, Y.; Shelhamer, E.; Donahue, J.; Karayev, S.; Long, J.; Girshick, R.;
  Guadarrama, S.; and Darrell, T.
\newblock 2014.
\newblock Caffe: Convolutional architecture for fast feature embedding.
\newblock {\em arXiv preprint arXiv:1408.5093}.

\bibitem[\protect\citeauthoryear{Krizhevsky, Sutskever, and
  Hinton}{2012}]{Krizhevsky2012ImageNet}
Krizhevsky, A.; Sutskever, I.; and Hinton, G.~E.
\newblock 2012.
\newblock Imagenet classification with deep convolutional neural networks.
\newblock In {\em NIPS},  1097--1105.

\bibitem[\protect\citeauthoryear{Li \bgroup et al\mbox.\egroup
  }{2016}]{li2016pruning}
Li, H.; Kadav, A.; Durdanovic, I.; Samet, H.; and Graf, H.~P.
\newblock 2016.
\newblock Pruning filters for efficient convnets.
\newblock {\em arXiv preprint arXiv:1608.08710}.

\bibitem[\protect\citeauthoryear{Li, Zhang, and Liu}{2016}]{li2016ternary}
Li, F.; Zhang, B.; and Liu, B.
\newblock 2016.
\newblock Ternary weight networks.
\newblock {\em arXiv preprint arXiv:1605.04711}.

\bibitem[\protect\citeauthoryear{Long, Shelhamer, and
  Darrell}{2015}]{Long_2015_CVPR}
Long, J.; Shelhamer, E.; and Darrell, T.
\newblock 2015.
\newblock Fully convolutional networks for semantic segmentation.
\newblock In {\em CVPR}.

\bibitem[\protect\citeauthoryear{Luo, Wu, and Lin}{2017}]{luo2017thinet}
Luo, J.-H.; Wu, J.; and Lin, W.
\newblock 2017.
\newblock Thinet: A filter level pruning method for deep neural network
  compression.
\newblock In {\em ICCV}.

\bibitem[\protect\citeauthoryear{Mellempudi \bgroup et al\mbox.\egroup
  }{2017}]{mellempudi2017ternary}
Mellempudi, N.; Kundu, A.; Mudigere, D.; Das, D.; Kaul, B.; and Dubey, P.
\newblock 2017.
\newblock Ternary neural networks with fine-grained quantization.
\newblock {\em arXiv preprint arXiv:1705.01462}.

\bibitem[\protect\citeauthoryear{Rastegari \bgroup et al\mbox.\egroup
  }{2016}]{rastegari2016xnor}
Rastegari, M.; Ordonez, V.; Redmon, J.; and Farhadi, A.
\newblock 2016.
\newblock Xnor-net: Imagenet classification using binary convolutional neural
  networks.
\newblock In {\em ECCV},  525--542.
\newblock Springer.

\bibitem[\protect\citeauthoryear{Ren \bgroup et al\mbox.\egroup
  }{2015}]{ren2015faster}
Ren, S.; He, K.; Girshick, R.; and Sun, J.
\newblock 2015.
\newblock Faster r-cnn: Towards real-time object detection with region proposal
  networks.
\newblock In {\em NIPS},  91--99.

\bibitem[\protect\citeauthoryear{Simonyan and
  Zisserman}{2014}]{simonyan2014very}
Simonyan, K., and Zisserman, A.
\newblock 2014.
\newblock Very deep convolutional networks for large-scale image recognition.
\newblock {\em arXiv preprint arXiv:1409.1556}.

\bibitem[\protect\citeauthoryear{Sun, Wang, and Tang}{2014}]{Sun_2014_CVPR}
Sun, Y.; Wang, X.; and Tang, X.
\newblock 2014.
\newblock Deep learning face representation from predicting 10,000 classes.
\newblock In {\em CVPR}.

\bibitem[\protect\citeauthoryear{Szegedy \bgroup et al\mbox.\egroup
  }{2015}]{szegedy2015going}
Szegedy, C.; Liu, W.; Jia, Y.; Sermanet, P.; Reed, S.; Anguelov, D.; Erhan, D.;
  Vanhoucke, V.; and Rabinovich, A.
\newblock 2015.
\newblock Going deeper with convolutions.
\newblock In {\em CVPR},  1--9.

\bibitem[\protect\citeauthoryear{Taigman \bgroup et al\mbox.\egroup
  }{2014}]{Taigman_2014_CVPR}
Taigman, Y.; Yang, M.; Ranzato, M.; and Wolf, L.
\newblock 2014.
\newblock Deepface: Closing the gap to human-level performance in face
  verification.
\newblock In {\em CVPR}.

\bibitem[\protect\citeauthoryear{Tang, Hua, and Wang}{2017}]{tang2017train}
Tang, W.; Hua, G.; and Wang, L.
\newblock 2017.
\newblock How to train a compact binary neural network with high accuracy?
\newblock In {\em AAAI},  2625--2631.

\bibitem[\protect\citeauthoryear{Yu \bgroup et al\mbox.\egroup
  }{2017}]{yu2017compressing}
Yu, X.; Liu, T.; Wang, X.; and Tao, D.
\newblock 2017.
\newblock On compressing deep models by low rank and sparse decomposition.
\newblock In {\em CVPR},  7370--7379.

\bibitem[\protect\citeauthoryear{Zhang \bgroup et al\mbox.\egroup
  }{2017}]{zhang2017shufflenet}
Zhang, X.; Zhou, X.; Lin, M.; and Sun, J.
\newblock 2017.
\newblock Shufflenet: An extremely efficient convolutional neural network for
  mobile devices.
\newblock {\em arXiv preprint arXiv:1707.01083}.

\bibitem[\protect\citeauthoryear{Zhou \bgroup et al\mbox.\egroup
  }{2016}]{zhou2016dorefa}
Zhou, S.; Wu, Y.; Ni, Z.; Zhou, X.; Wen, H.; and Zou, Y.
\newblock 2016.
\newblock Dorefa-net: Training low bitwidth convolutional neural networks with
  low bitwidth gradients.
\newblock {\em arXiv preprint arXiv:1606.06160}.

\bibitem[\protect\citeauthoryear{Zhou \bgroup et al\mbox.\egroup
  }{2017}]{zhou2017incremental}
Zhou, A.; Yao, A.; Guo, Y.; Xu, L.; and Chen, Y.
\newblock 2017.
\newblock Incremental network quantization: Towards lossless cnns with
  low-precision weights.
\newblock {\em arXiv preprint arXiv:1702.03044}.

\bibitem[\protect\citeauthoryear{Zhu \bgroup et al\mbox.\egroup
  }{2016}]{zhu2016trained}
Zhu, C.; Han, S.; Mao, H.; and Dally, W.~J.
\newblock 2016.
\newblock Trained ternary quantization.
\newblock {\em arXiv preprint arXiv:1612.01064}.

\end{thebibliography}

\end{document}